\newcommand{\figdir}{}

\documentclass[times, 10pt,twocolumn]{article}

\usepackage{paper}
\usepackage{times}
\usepackage{balance}

\usepackage{epsfig}
\usepackage{graphicx}
\usepackage{amssymb,amsmath}

\begin{document}

\title{Convolutional Neural Networks Applied to\\
  House Numbers Digit Classification}

\author{Pierre Sermanet, Soumith Chintala and Yann LeCun\\
\emph{The Courant Institute of Mathematical Sciences - New York University}\\
\emph{\{sermanet,soumith,yann\}@cs.nyu.edu}\\
%\and
%First line of institution2 address\\ Second line of institution2 address\\
%SecondAuthor@institution2.com\\
}

\maketitle
\thispagestyle{empty}

\begin{abstract}
We classify digits of real-world house numbers using convolutional neural networks (ConvNets). ConvNets are hierarchical feature learning neural networks whose structure is biologically inspired. Unlike many popular vision approaches that are hand-designed, ConvNets can automatically learn a unique set of features optimized for a given task. We augmented the traditional ConvNet architecture by learning
multi-stage features and by using Lp pooling and establish a new
state-of-the-art of $94.85\%$ accuracy on the SVHN dataset ($45.2\%$ error improvement). Furthermore, we analyze the benefits of different pooling methods and multi-stage features in ConvNets. The source code and a tutorial are available at eblearn.sf.net.
\end{abstract}

%-------------------------------------------------------------------------
\vspace*{-.2cm}
\Section{Introduction}

\begin{figure}[h]
  \begin{center}
  \begin{minipage}[t]{0.5\textwidth}
    \vspace{0pt}
    \includegraphics[width=0.9\textwidth]{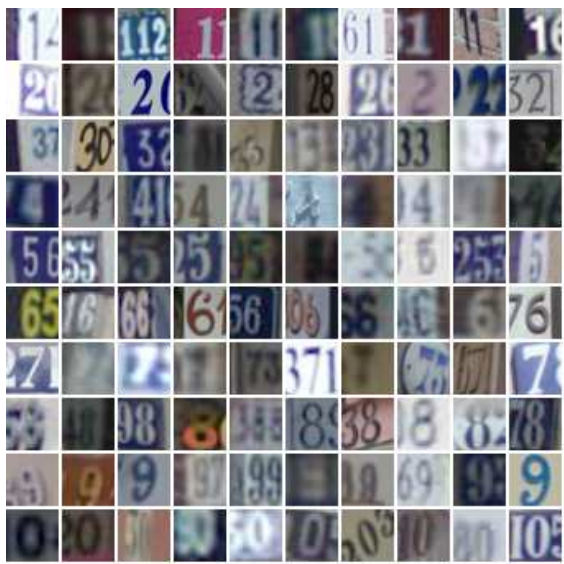}
  \end{minipage}
  \end{center}
  \caption{\small{32x32 cropped samples from the classification task of the SVHN dataset. Each sample is assigned only a single digit label (0 to 9) corresponding to the center digit.
  }}
  \label{fig:fig1}
\vspace*{-.5cm}
\end{figure}

Character recognition in documents can be considered a solved task for computer
vision, whether handwritten or typed. It is however a harder problem in the
context of complex natural scenes like photographs where the best current
methods lag behind human performance, mainly due to non-contrasting
backgrounds, low resolution, de-focused and motion-blurred images and large
illumination differences (Figure~\ref{fig:fig1}). 

~\cite{netzer-11} recently introduced a new digit classification dataset of house numbers extracted from street level images. It is similar in format to the popular MNIST dataset\cite{mnistlecun} (10 digits, 32x32 inputs), but an order of magnitude bigger (600,000 labeled digits), contains color information and various natural backgrounds.

Previous approaches in classifying characters and digits from natural images used multiple hand-crafted features~\cite{deCampos09} and template-matching~\cite{YamaguchiNMMH03}. In contrast, ConvNets learn features all the way from pixels to the classifier.~\cite{netzer-11} demonstrated the superiority of learned features over hand-designed ones. Such superiority was also previously shown among others in a traffic sign classification challenge~\cite{gtsrb-11} where two independent teams obtained the best performance against various other approaches using ConvNets~\cite{sermanet-ijcnn-11,ciresan-11}. ~\cite{netzer-11} also show superior results with unsupervised learning, we however only report results with fully-supervised training. We obtain a $4.25$ points improvement in accuracy (with
$94.85\%$ accuracy) over the previous state-of-the-art of $90.6\%$. 
We use the traditional ConvNet architecture augmented with different pooling methods and with multi-stage features~\cite{sermanet-ijcnn-11}. This work was
implemented with the EBLearn~\footnote{http://eblearn.sf.net}
C++ open-source framework~\cite{sermanet-ictai-09}.

%-------------------------------------------------------------------------
\vspace*{-.2cm}
\section{Architecture}
\vspace*{-.2cm}
The ConvNet architecture is composed of repeatedly stacked feature stages.
Each stage contains a convolution module, followed by a pooling/subsampling module and a normalization module. While traditional pooling modules in ConvNet are either average or max poolings, we use an Lp pooling here. The normalization module is subtractive only as opposed to subtractive and divisive, i.e. the mean value of each neighborhood is subtracted to the output of each stage (but not divided by the standard deviation as it decreases performance with this dataset).
Finally, multi-stage features are also used as opposed to single-stage features.

\subsection{Lp-Pooling}
\vspace*{-.5cm}

\begin{figure}[h]
  \begin{center}
    \includegraphics[width=.5\textwidth]{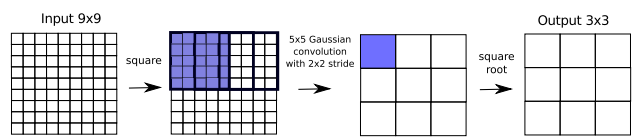}
  \end{center}
  \caption{\small{L2-pooling applied to a 9x9 feature map with a 3x3 Gaussian kernel and 2x2 stride}}
  \label{fig:fig2}
\end{figure}
\vspace*{-.5cm}

Lp pooling is a biologically inspired pooling layer modelled on complex cells~\cite{Simoncelli97amodel,Hyvarinen05complexcell} who's operation can be summarized in equation (1), where $G$ is a Gaussian kernel, $I$ is the input feature map and $O$ is the output feature map. It can be imagined as giving an increased weight to stronger features and suppressing weaker features. Two special cases of Lp pooling are notable. $P = 1$ corresponds to a simple Gaussian averaging, whereas $P = \infty$ corresponds to max-pooling (i.e only the strongest signal is activated). Lp-pooling has been used previously in~\cite{koray-cvpr-09,Yang09linearspatial} and a theoretical analysis of this method is described in~\cite{boureau-icml-10}.
\begin{equation}
O = (\sum \sum I(i,j)^P \times G(i,j))^{1/P} 
\end{equation}

Figure~\ref{fig:fig2} demonstrates a simple example of L2-pooling.

\begin{table*}[htb] 
  \small{
    \begin{center}
      \begin{tabular}{ | c | c | c | c |}
        \hline
        Task & Single-Stage features & Multi-Stage features & Improvement \% \\
        \hline \hline
        Pedestrians detection (INRIA)~\cite{sermanet-snowbird-11}& 14.26\% & 9.85\% & 31\%\\
        Traffic Signs classification (GTSRB)~\cite{sermanet-ijcnn-11}& 1.80\% & 0.83\% & 54\%\\
        House Numbers classification (SVHN) & 5.72\% & 5.67\% & 0.9\%\\
        \hline
      \end{tabular}
    \end{center}
    \caption{\small{Error rates improvements of multi-stage features over
        single-stage features for different types of objects detection and
        classification. Improvements are significant for multi-scale and
        textured objects such as traffic signs and pedestrians but minimal for
        house numbers.}}
  }
  \label{table:ms}
\vspace*{-1.5cm}
\end{table*}

\begin{figure}[h]
  \begin{center}
    \begin{minipage}[t]{0.5\textwidth}
    \vspace{0pt}
      \includegraphics[width=1\textwidth]{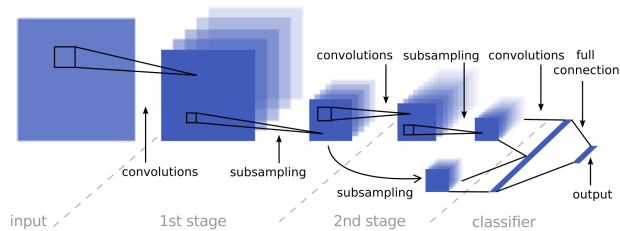}
  \end{minipage}
  \end{center} 
  \caption{\small{A 2-stage ConvNet architecture where Multi-Stage features
      (MS) are fed to a 2-layer classifier. The 1st stage features are branched	
      out, subsampled again and then concatenated to 2nd stage features.}}
  \label{fig:arch}
\vspace*{-.5cm}
\end{figure}

\subsection{Multi-Stage Features}

\begin{figure}[h]
  \begin{center}
    \begin{minipage}[t]{0.5\textwidth}
    \vspace{0pt}
      \includegraphics[width=1\textwidth]{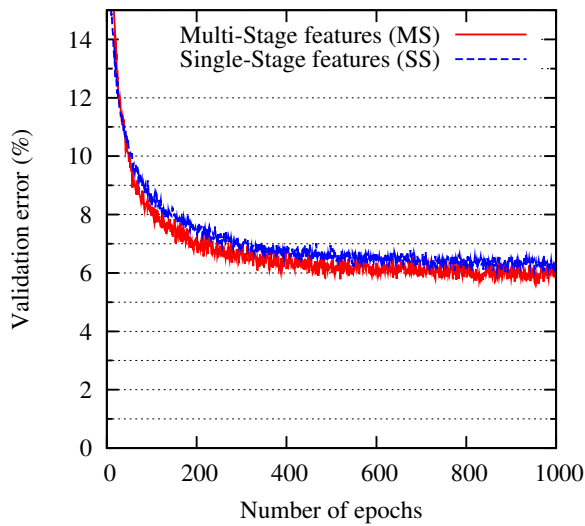}
  \end{minipage}
  \end{center} 
  \caption{\small{Improvement of Multi-Stage features (MS) over Single-Stage features (SS) in error rate on the validation set. MS features provide a slight
      error improvement over SS features.}}
  \label{fig:ms}
\end{figure}

Multi-Stage features (MS) are obtained by branching out outputs of all stages into the classifier (Figure~\ref{fig:arch}). They provide richer representations compared to Single-Stage features (SS) by adding complementary information such as local textures and fine details lost by higher levels. MS features have consistently improved performance in other work~\cite{fan-nn-10,sermanet-ijcnn-11,sermanet-snowbird-11} and in this work as well (Figure~\ref{fig:ms}). However we observe minimal gains on this dataset compared to other types of objects such as pedestrians and traffic signs (Table 1). The likely explanation for this observation is that gains are correlated to the amount of texture and multi-scale characteristics of the objects of interest.

%-------------------------------------------------------------------------
\vspace*{-.2cm}
\Section{Experiments}
\vspace*{-.2cm}

\SubSection{Data Preparation}
The SVHN classification dataset~\cite{netzer-11} contains 32x32 images
with 3 color channels. The dataset is divided into three subsets: train set, extra set and test set. The extra set is a large set of easy samples and train set is a smaller set of more difficult samples. Since we are given no information about how the sampling of these images was done, we assume a random order to construct our validation set. We compose our validation set with $2/3$ from training samples (400 per class) and $1/3$ from extra samples (200 per class), yielding a total of 6000 samples. This distribution allows to measure success on easy samples but puts more emphasis on difficult ones.

Samples are pre-processed with a local contrast normalization (with a 7x7 kernel) on the Y channel of the YUV space followed by a global contrast normalization over each channel. No sample distortions were used to improve invariance.

%\vspace*{-1cm}
%\vskip -16pt

%\vskip -16pt
 
\subsection{Architecture Details} 
The ConvNet has 2 stages of feature extraction and a two-layer non-linear classifier. The first convolution layer produces 16 features with 5x5 convolution filters while the second convolution layer outputs 512 features with 7x7 filters. The output to the classifier also includes inputs from the first layer, which provides local features/motifs to reinforce the global features. The classifier is a 2-layer non-linear classifier with 20 hidden units. 
Hyper-parameters such as learning rate, regularization constant and learning rate decay were tuned on the validation set.
We use stochastic gradient descent as our optimization method and shuffle our dataset after each training iteration.

\begin{figure}[h]
  \begin{center}
    \begin{minipage}[t]{0.5\textwidth}
    \vspace{0pt}
      \includegraphics[width=1\textwidth]{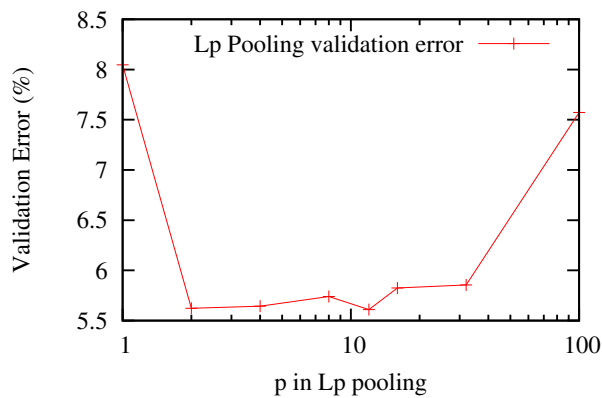}
  \end{minipage}
  \end{center} 
  \caption{\small{Error rate of Lp-pooling on the validation set for
      $p = 1,2,4,8,12,16,32,\infty$ ($p=\infty$} is represented as $p=100$ for
    convenience). These validation errors are reported after 1000 training
    epochs. $p = 12$ performs best with an error rate of $5.61\%$.}
  \label{fig:fig3}
\vspace*{-.5cm}
\end{figure}

For the pooling layers, we compare Lp-pooling for the value $p = 1,2,4,8,12,16,32,\infty$ on the validation set and use the best performing pooling on the final testing. The performance of different pooling methods on the validation set can be seen in Figure~\ref{fig:fig3}. Insights from~\cite{boureau-icml-10} tell us that the optimal value of $p$ varies for different input spaces and there is no single globally optimal value for $p$. For our validation data, we observe that $p = 2,4,12$ give the best performance ($5.62\%, 5.64\%$ and $5.61\%$ respectively). Max-pooling, which corresponds to $p = \infty$ yielded a validation error rate of $7.57\%$.

\begin{table*}[htb]
  \small{
  \begin{center}
    \begin{tabular}{ | c | c | }
      \hline
      Algorithm & SVHN-Test Accuracy \\
      \hline \hline
      Binary Features (WDCH) & 63.3\% \\
      HOG & 85.0\% \\
      Stacked Sparse Auto-Encoders & 89.7 \% \\
      K-Means & 90.6\% \\
      {\bf ConvNet / MS / Average} & {\bf 90.75\%} \\
      {\bf ConvNet / MS / L2 / Smaller training} & {\bf 91.55\%} \\
      {\bf ConvNet / SS / L2} & {\bf 94.28\%} \\
      {\bf ConvNet / MS / L2} & {\bf 94.33\%} \\
      {\bf ConvNet / MS / L12} & {\bf 94.76\%} \\
      {\bf ConvNet / MS / L4} & {\bf 94.85\%} \\
      %% {\bf ConvNet w/ max pooling} & {\bf 84.45\%} \\
      \hline
      Human Performance &98.0\% \\
      \hline
    \end{tabular}
  \label{table:res}
  \end{center}
\vspace*{-.2cm}
  \caption{\small{Performance reported by~\cite{netzer-11} with the additional
      Supervised ConvNet with state-of-the-art accuracy of 94.85\%.}}
  }
\end{table*}

\section{Results \& Future Work}

Our experiments demonstrate a clear advantage of Lp pooling with $1 < p < \infty$ on this dataset in validation (Figure~\ref{fig:fig3}) and test (Average pooling is 3.58 points inferior to L2 pooling in Table 2). With L4 pooling, we obtain a state-of-the-art performance on the test set with an accuracy of 94.85\% compared to the previous best of 90.6\% (Table 2). We also show that using multi-stage features gives only a slight increase in performance, compared to the performance increase seen in other vision applications.

\begin{figure*}[htb]
  \begin{center}
      \includegraphics[width=.9\textwidth]{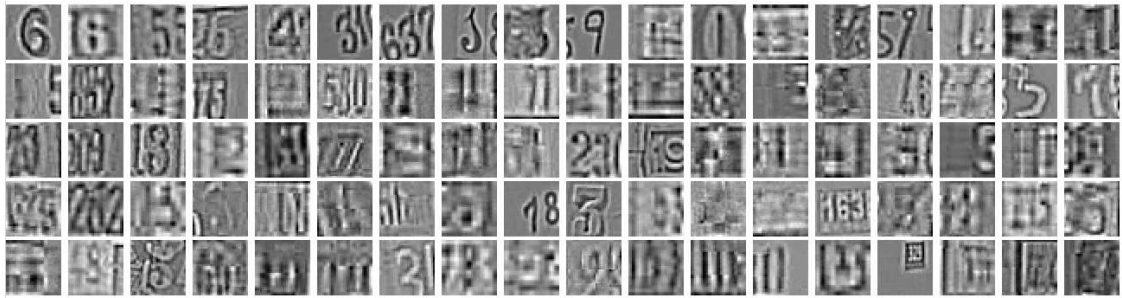}
  \end{center} 
\vspace*{-.5cm}
  \caption{\small{Preprocessed Y channel of validation samples with
      highest energy (i.e. highest error)
      with the 94.33\% accuracy L2-pool based multi-stage ConvNet.}}
  \label{fig:highest}
\vspace*{-.2cm}
\end{figure*}

Additionally, it is important to note that our approach is trained fully supervised only, whereas the best previous methods are unsupervised learning methods (k-means, auto-encoders). We shall, in the future, run experiments with unsupervised learning, to compare the accuracy improvement that can be attributed to supervision. Figure~\ref{fig:highest} shows the validation samples with highest energy. Many of these seem to exhibit large scale variations, future work could address this problem by introducing artificial scale deformations during training.

\balance

%-------------------------------------------------------------------------
%\nocite{ex1,ex2}
\bibliographystyle{latex12}
%\small{
\bibliography{paper}

\begin{thebibliography}{10}\setlength{\itemsep}{-1ex}\small

\bibitem{boureau-icml-10}
Y.~Boureau, J.~Ponce, and Y.~LeCun.
\newblock A theoretical analysis of feature pooling in vision algorithms.
\newblock In {\em Proc. International Conference on Machine learning}, 2010.

\bibitem{ciresan-11}
D.~C. Ciresan, U.~Meier, J.~Masci, and J.~Schmidhuber.
\newblock A committee of neural networks for traffic sign classification.
\newblock In {\em International Joint Conference on Neural Networks}, pages
  1918--1921, 2011.

\bibitem{deCampos09}
T.~E. de~Campos, B.~R. Babu, and M.~Varma.
\newblock Character recognition in natural images.
\newblock In {\em Proceedings of the International Conference on Computer
  Vision Theory and Applications, Lisbon, Portugal}, February 2009.

\bibitem{fan-nn-10}
J.~Fan, W.~Xu, Y.~Wu, and Y.~Gong.
\newblock Human tracking using convolutional neural networks.
\newblock {\em Neural Networks, IEEE Transactions on}, 21(10):1610 --1623,
  2010.

\bibitem{Hyvarinen05complexcell}
A.~Hyvärinen and U.~Köster.
\newblock Complex cell pooling and the statistics of natural images.
\newblock In {\em Computation in Neural Systems,}, 2005.

\bibitem{koray-cvpr-09}
K.~Kavukcuoglu, M.~Ranzato, R.~Fergus, and Y.~LeCun.
\newblock Learning invariant features through topographic filter maps.
\newblock In {\em Proc. International Conference on Computer Vision and Pattern
  Recognition}. IEEE, 2009.

\bibitem{mnistlecun}
Y.~Lecun and C.~Cortes.
\newblock {The MNIST database of handwritten digits}.

\bibitem{netzer-11}
Y.~Netzer, T.~Wang, A.~Coates, A.~Bissacco, B.~Wu, and A.~Y. Ng.
\newblock Reading digits in natural images with unsupervised feature learning.
\newblock In {\em NIPS Workshop on Deep Learning and Unsupervised Feature
  Learning}, 2011.

\bibitem{sermanet-snowbird-11}
P.~Sermanet, K.~Kavukcuoglu, and Y.~LeCun.
\newblock Traffic signs and pedestrians vision with multi-scale convolutional
  networks.
\newblock In {\em Snowbird Machine Learning Workshop, 2011}.

\bibitem{sermanet-ictai-09}
P.~Sermanet, K.~Kavukcuoglu, and Y.~LeCun.
\newblock Eblearn: Open-source energy-based learning in c++.
\newblock In {\em Proc. International Conference on Tools with Artificial
  Intelligence}. IEEE, 2009.

\bibitem{sermanet-ijcnn-11}
P.~Sermanet and Y.~LeCun.
\newblock Traffic sign recognition with multi-scale convolutional networks.
\newblock In {\em Proceedings of International Joint Conference on Neural
  Networks}, 2011.

\bibitem{Simoncelli97amodel}
E.~P. Simoncelli and D.~J. Heeger.
\newblock A model of neuronal responses in visual area mt, 1997.

\bibitem{gtsrb-11}
J.~Stallkamp, M.~Schlipsing, J.~Salmen, and C.~Igel.
\newblock The {G}erman {T}raffic {S}ign {R}ecognition {B}enchmark: A
  multi-class classification competition.
\newblock In {\em IEEE International Joint Conference on Neural Networks},
  pages 1453--1460, 2011.

\bibitem{YamaguchiNMMH03}
T.~Yamaguchi, Y.~Nakano, M.~Maruyama, H.~Miyao, and T.~Hananoi.
\newblock Digit classification on signboards for telephone number recognition.
\newblock In {\em ICDAR}, pages 359--363, 2003.

\bibitem{Yang09linearspatial}
J.~Yang, K.~Yu, Y.~Gong, and T.~Huang.
\newblock Linear spatial pyramid matching using sparse coding for image
  classification.
\newblock In {\em in IEEE Conference on Computer Vision and Pattern
  Recognition}, 2009.

\end{thebibliography}
%}
\end{document}